\documentclass{article}

\usepackage{arxiv}

\usepackage[utf8]{inputenc} 
\usepackage[T1]{fontenc}    
\usepackage{hyperref}       
\usepackage{url}            
\usepackage{booktabs}       
\usepackage{amsfonts}       
\usepackage{nicefrac}       
\usepackage{microtype}      
\usepackage{lipsum}		
\usepackage{graphicx}
\usepackage{doi}
\usepackage[perpage]{footmisc}
\usepackage{multirow}

\usepackage[inline]{enumitem}
\usepackage{amsmath}
\usepackage{cleveref}
\usepackage{float}
\usepackage[numbers]{natbib}

\title{PredictChain: Empowering Collaboration and Data Accessibility for AI in a Decentralized Blockchain-based Marketplace}

\author{ 
    {Matthew T. Pisano} \\
	Department of Computer Science\\
	Rensselaer Polytechnic Institute\\
	Troy, NY, USA \\
	\And
	{Connor J. Patterson} \\
	Department of Computer Science\\
	Rensselaer Polytechnic Institute\\
	Troy, NY, USA \\
	\And
	{Oshani Seneviratne} \\
	Department of Computer Science\\
	Rensselaer Polytechnic Institute\\
	Troy, NY, USA \\
}

\date{}

\renewcommand{\headeright}{}
\renewcommand{\undertitle}{}
\renewcommand{\shorttitle}{PredictChain}

\hypersetup{
pdftitle={PredictChain},
pdfsubject={Blockchain-based AI Marketplace},
pdfauthor={Matthew T. Pisano,  Connor J. Patterson and Oshani Seneviratne},
pdfkeywords={Blockchain, Decentralized AI and Data Marketplace, Oracle, Algorand},
}

\begin{document}

\maketitle

\begin{abstract}
Limited access to computing resources and training data poses significant challenges for individuals and groups aiming to train and utilize predictive machine learning models. Although numerous publicly available machine learning models exist, they are often unhosted, necessitating end-users to establish their computational infrastructure. Alternatively, these models may only be accessible through paid cloud-based mechanisms, which can prove costly for general public utilization. Moreover, model and data providers require a more streamlined approach to track resource usage and capitalize on subsequent usage by others, both financially and otherwise. An effective mechanism is also lacking to contribute high-quality data for improving model performance.
We propose a blockchain-based marketplace called \emph{``PredictChain''} for predictive machine-learning models to address these issues. This marketplace enables users to upload datasets for training predictive machine learning models, request model training on previously uploaded datasets, or submit queries to trained models. Nodes within the blockchain network, equipped with available computing resources, will operate these models, offering a range of archetype machine learning models with varying characteristics, such as cost, speed, simplicity, power, and cost-effectiveness. This decentralized approach empowers users to develop improved models accessible to the public, promotes data sharing, and reduces reliance on centralized cloud providers.


\end{abstract}


\section{Introduction}

AI is becoming pervasive in our lives, from voice assistants to self-driving cars~\cite{whittaker2018ai}. However, developing and training many machine learning models requires enormous data and computing resources. For instance, state-of-the-art image classification models like ResNet-50~\cite{tai2017image} are designed to be trained on high-end GPUs on the ImageNet dataset~\cite{deng2009imagenet}, which has over 14 million hand-annotated images. Such resource requirements have led to the concentration of AI development, which creates an obstacle to innovation by the masses.

Decentralized AI marketplaces (such as SingularityNET~\cite{montes2019distributed} and Ocean Protocol~\cite{mcconaghy2022ocean}) have emerged as promising solutions to these challenges, which allow individuals and organizations to collectively share their data and computational resources to develop and train machine learning models. By leveraging blockchain technology, decentralized AI marketplaces can facilitate secure, transparent, incentivized, and decentralized access to machine learning models and data, promoting innovation and collaboration. 

In this paper, we propose a blockchain-based marketplace for AI called ``PredictChain'' that aims to democratize machine learning model training and inference using the Algorand blockchain on several machine learning models.
When users upload their datasets to PredictChain, they allow a model to be trained on the datasets uploaded. Higher-quality datasets will produce higher-quality models. The model owners and the dataset contributors are rewarded for their contributions when a model is queried. 
The amount of this reward is based on the correctness of the prediction, which encourages users to participate in contributing the resources needed for good predictions while leaving a public record for other users to view, which is in contrast to large cloud organizations, such as \textit{AWS} or \textit{Google Cloud}. These cloud providers allow users to train models on custom datasets, and come at a price, with no chance of reward if a model does well.

\section{Related Work}

The development of decentralized AI marketplaces using blockchain technology is an evolving field that has gained significant attention recently. While the concept of AI resource marketplaces is not new, utilizing blockchain technology enables the creation of decentralized platforms between untrusted parties, potentially transforming AI model training and development.
Various studies have investigated the application of blockchain technology in establishing decentralized marketplaces for AI resources, as we outline below. 



Harris and Waggoner outline a blockchain-based predictive marketplace that utilizes AI for predictive analysis. In their solution, an initial model is trained on the data collaboratively provided by multiple users~\cite{sharingModels,harris2020leveraging,harris2020analysis}. To encourage the submission of high-quality data, the authors propose an incentive mechanism that rewards honest users and punishes dishonest users. Their model is trained off-chain on the client's machine. All the data submission and incentives logic is done within an Ethereum smart contract.
%
This system leaves the exact definition of the model flexible between different clients running the system. They suggest that the specific class of model would be some predictive model, giving examples of Conditional Random Fields as Recurrent Neural Networks (CRF-RNNs) and Support Vector Machines (SVMs). Depending on the model used, the authors mention that the corresponding incentive mechanism would change from model to model.
%
The incentives proposed in this work fall into gamification and monetary incentives. In the gamification strategy, the authors suggest that there should be no monetary incentive for contributing data, instead using tokens or badges in a game-theoretic manner. For the monetary route, they suggest that users who upload data may also have to submit some stake into the contract as collateral.
Our work is most similar to this but with some notable differences. We have deployed our solution on the Algorand blockchain, which provides several performance advantages over an Ethereum-based solution, and we have leveraged some advanced machine learning models in our solution offering.


Sarpatwar et al. take some of the decentralization ideas of the previous work and expand upon them by structuring their project with security as a main focus~\cite{priceOfTrust}. They develop a marketplace for health insurance entities needing models trained on confidential data, and the primary features of their work are \textit{trust}, \textit{fairness}, and \textit{auditability}.
%
The datasets they used are unique since the bulk of the data is not tied directly to any one training request.
For instance, a user uploads only a small chunk of data. This data will then be used as the validation set for the training. Most of the training data for the model inference process is pulled from private sources, requiring users to upload only a small portion of their data. However, this has the disadvantage of requiring external data to train models. To accommodate this, the authors implemented several layers of security to protect private data sources.
%
The models they target need to be more integrated and trained in a decentralized manner. This training strategy reflects federated learning~\cite{yang2019federated}, where each model is given only some data in bulk and only sections with confidential information stripped out.
Additionally, the models are eventually shared with the requesting user along with the model results.
%
An off-chain component communicates with this system's private, permissioned blockchain (Hyperledger Fabric~\cite{androulaki2018hyperledger}), which allows for more complexities in training the models, as on-chain training has significant speed and memory constraints.

Montes and Goertzel have also noticed the problem of accessibility in the realm of AI and have proposed a solution using blockchain technologies~\cite{democratizedAI}. Their marketplace implementation, \textit{SingularityNET}, was designed to make AI development and training more accessible to everyone. They argue that this democratization is useful for allowing more people to access AI technologies, and it significantly impacts the development of artificial general intelligence (AGI).
%
Furthermore, they also note that the current development of AI systems is controlled primarily by only a handful of large tech giants and that this degree of centralization would hurt any eventual AGI due to the bias that it may be given when trained only by a few thousand individuals, missing out on important data that can be gathered from outside these organizations.
%
Their solution to this centralization problem is the \textit{SingularityNET} marketplace. There, users can request or provide AI services. These actions are organized using simple, smart contracts that take in payment from the requesting users and distribute rewards to service-providing users. Utilizing blockchain and smart contract technologies is designed to avoid the closed and centralized qualities that many contemporary AI marketplaces embody.


Leveraging blockchain to enhance the security and privacy of machine learning models in specific domains has also been explored.

Shamsi and Cuffe demonstrate a market similar to ours but using a more decentralized architecture and a more specific scope for a use case of blockchain-empowered prediction markets for the renewable energy market~\cite{windForcasting}. 
%
A key problem with many renewable energy sources is their unpredictable supply. Wind turbines need wind. Solar panels need sunshine. Forecasting the output of these energy production methods at various times or places helps alleviate the pitfalls that prevent them from growing in market share.
%
On their platform, anyone can propose an event and request predictions for that event with the promise of providing a reward. Reporters then propose several predictions throughout the seven-day window that the market is active. In this implementation, the hosts of this predictive market do not take responsibility for the training and execution of the predictive machine learning models; instead, it is all done on the side of the reporters. Using their framework, reporters who provide correct data are rewarded, and those who provide inconsistent data are punished.

%
Shen and Shafiq focus on using big data to train predictive machine learning models to forecast the short-term future behavior of stocks within the Chinese stock market~\cite{deepPrediction}. For their dataset, they gathered the past performance of 3558 stocks from the Chinese stock market for two years. 
To gather this data, they utilized the \textit{Tushare} API\footnote{\url{https://pypi.org/project/tushare}}, along with web-scraping from \textit{Sina Finance}\footnote{\url{https://finance.sina.cn}} and the \textit{SWS Research}\footnote{\url{https://www.swsresearch.com/institute_sw/home}} websites. 
The authors utilized several techniques to reduce the noise inherent to the stock market to allow for better predictions. Their primary method for doing this was through feature engineering, for which they utilized three primary techniques. First, they applied feature extension to their dataset. Using this technique, they added additional meta-attributes to the entries, such as polarizing or calculating the fluctuation percentage. Adding this metadata to the dataset helps to give the model a more complete picture of the stock and how it performs over time. Next, they eliminated some features, based on their influence, by using the Recursive Feature Elimination (RFE) algorithm. Eliminating unnecessary features allows their model to pay more attention to the most influential features, improving its predictive capabilities.
%
For testing, the authors used a two-layered Long Short Term Memory (LSTM) model with only an input layer and an output layer, and as for their output, they kept it as simple as possible to ensure the model was focused only on predicting the movement of the stock. The model either outputted a $1$ for the stock going up at a given time step or a $0$ if the model believed the stock would go down.
This work helped us better understand some of the future improvements we could make to our dataset preprocessing and model training. 

\section{Methodology}

We have expanded upon the ideas mentioned above and addressed some of their shortcomings in PredictChain, which presents a more scalable architecture using off-chain oracles and the Algorand blockchain that supports the instant finality of transactions. Our aim with this strategy is to focus on simplicity and ease of use. Our implementation allows anyone to quickly set up an oracle instance and begin connecting to clients without requiring a large network of oracles. Another design choice that we made was in our suite of built-in archetype machine learning models. Each client has a preset list of models to choose from. It is no longer the client's responsibility to train and implement the models, as in Harris and Waggoner (2019)~\cite{sharingModels}, which helps to make setup and usage much cleaner and more accessible. Thus, PredictChain aims to be more generalized and accessible to non-technical users.

\begin{figure}[!htbp]
\centering
\begin{minipage}{.5\textwidth}
  \centering
   \includegraphics[width=0.98\linewidth]{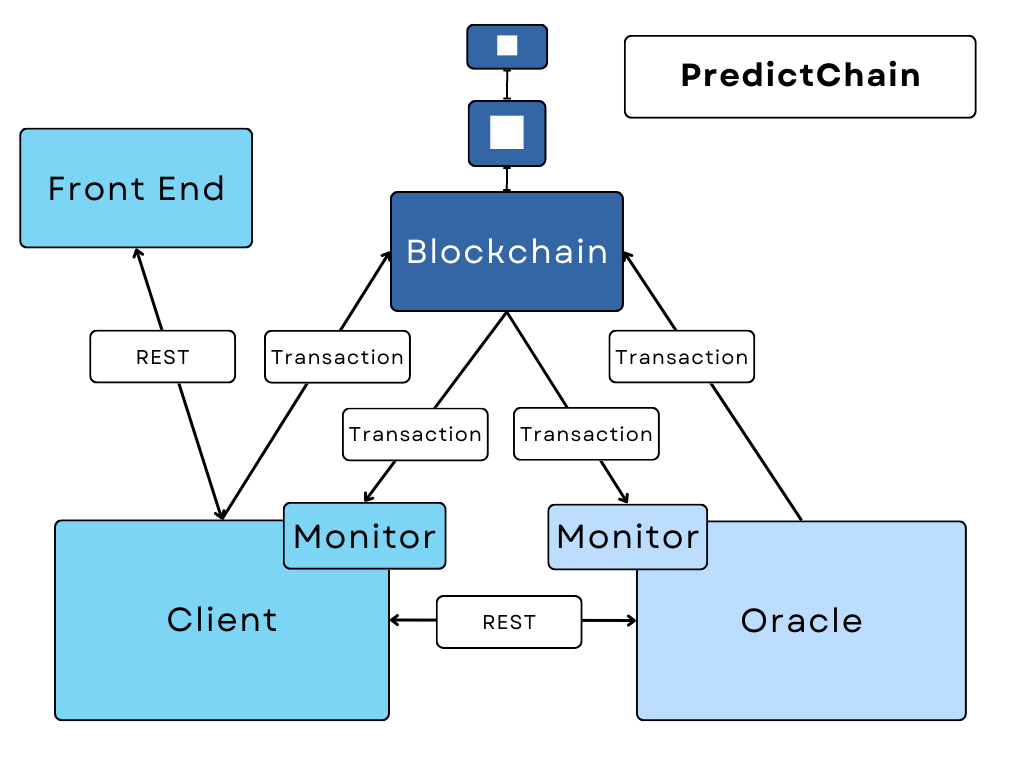}
   \caption{PredictChain Overall Architecture}
   \label{Fig:detailedDiagram}
   
\end{minipage}%
\begin{minipage}{.5\textwidth}
  \centering
  \includegraphics[width=0.98\linewidth]{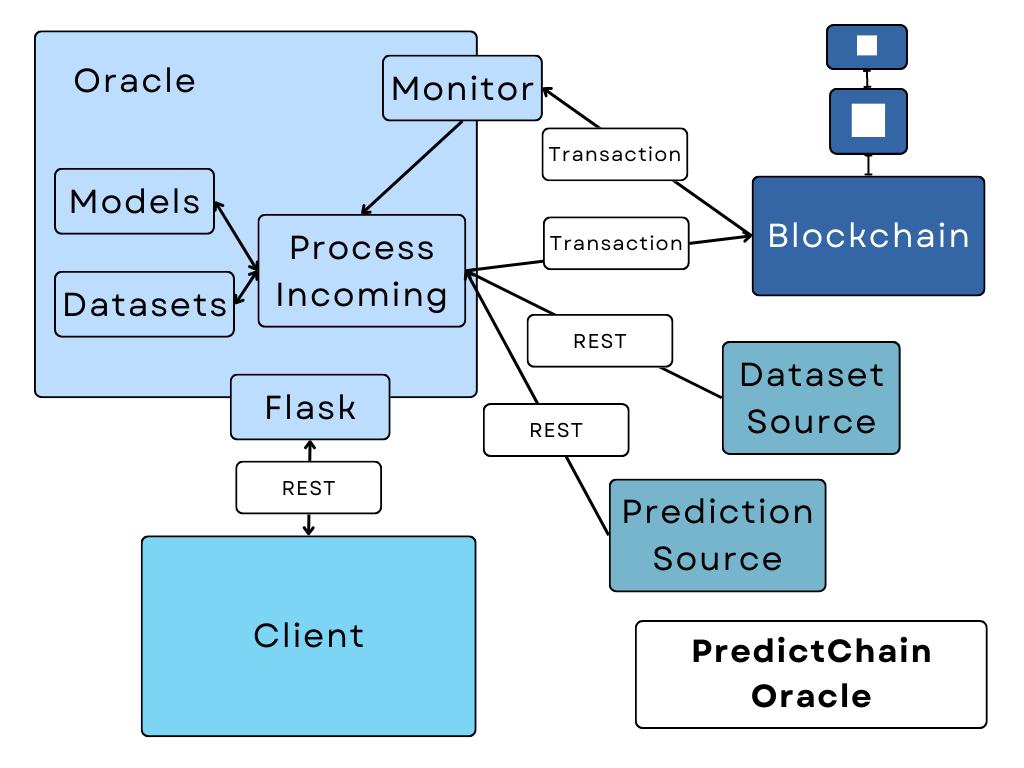}
   \caption{PredictChain Oracle}
   \label{Fig:oracleDiagram}
\end{minipage}
\end{figure}

Our solution consists of several key components to enable the sharing and utilization of AI resources. These components include a blockchain, an oracle, tokenomics, and the client. 
\Cref{Fig:detailedDiagram} illustrates the system's core components, along with their methods of communication.

\subsection{Blockchain}

Blockchain is useful in its role due to its immutability and its transparency. Using the Algorand blockchain means that all requests are permanently stored and public, so other users can see what type of models are useful for specific datasets and what results (predictions, classifications) those models have produced.
In PredictChain, blockchain serves as a records keeper and a messenger between the client and the oracle, accomplished by using transactions as a form of direct communication. With every transaction sent, there is a note. This note is a JSON-encoded string (encoded in base64) that communicates information about the operation that the transaction is requesting and arguments for that operation. A series of opcodes represent these operations. These codes are abbreviations of the operation names enclosed in angle brackets, for example, $\langle$\textit{QUERY\_MODEL}$\rangle$. The arguments to these operations are represented as a \emph{named dictionary}, with each key being the name of the argument and each value being the argument itself. This named strategy allows the program to be very flexible and agnostic about the exact ordering of the arguments. 

We utilize the Algorand Blockchain~\cite{gilad2017algorand} to implement PredictChain.
The Algorand blockchain is a decentralized, permissionless, and high-performing blockchain well-suited for creating a decentralized AI marketplace. First, Algorand's unique Pure Proof of Stake (PPoS) consensus algorithm~\cite{dimitri2022proof} and architecture provides several advantages that make it an ideal choice for this type of application. The PPoS consensus algorithm ensures fast and secure transaction processing, enabling efficient transactions within the marketplace, which means that participants can quickly and easily exchange resources without needing a centralized intermediary, reducing transaction costs and promoting decentralization.
Second, Algorand's smart contract capabilities allow for the creation of smart contracts that can be used to govern transactions within the marketplace. Smart contracts can be used to enforce rules and regulations around the use of data and computing resources, ensuring that participants' privacy and security are protected.
Third, Algorand's Layer-1 architecture enables scalable and high-performing applications. The platform can support many transactions per second, enabling participants to quickly and efficiently exchange resources within the marketplace.
Finally, Algorand's commitment to sustainability and energy efficiency aligns with the goals of creating more environmentally friendly and sustainable machine learning models. The platform's low energy consumption and carbon footprint make it an attractive choice for those seeking to reduce the environmental impact of AI development and training.
Algorand's unique features and architecture make it an ideal choice for this application, enabling greater collaboration and innovation.
Furthermore, using ALGOs as a payment method demonstrates that cryptocurrencies can effectively be used as a direct form of payment for useful services.

\subsection{Oracle}

The oracle (\Cref{Fig:oracleDiagram}) acts as the bridge between the off-chain world and the underlying blockchain, facilitating the retrieval and verification of external data required for training and prediction. 
It constantly polls for updates coming from the client through the blockchain by using its monitor.
Upon receiving these updates, it begins the execution of one of its three main operations:
\begin{enumerate*}
    \item Downloading a user-specified dataset and saving it.
    \item Training one of the raw models based on user-inputted parameters.
    \item Querying one of the trained models on user-inputted data and comparing it to the real-world result.
\end{enumerate*}
After each of these operations, the oracle sends out several blockchain transactions. These can be either rewards to contributors of a model or confirmations/results of the operation that has been performed.

When working with user-submitted datasets, the oracle uses a handler to manage the operations performed on that dataset. The handler can save datasets to a specified environment, load datasets from a specified environment, parse them as a data frame, and split the dataset by the values of one of its attributes. The environments that the handler recognizes are \textit{local} and \textit{remote} via the InterPlanetary File System (IPFS)~\cite{benet2014ipfs}. When using either of these environments, the handler abstracts away the complexities of working with them into a unified interface.

The oracle uses a similar, common interface when working with user-trained models. This interface can create the model architecture, train the model on a selected dataset, query the trained model, evaluate its performance, save the model, and load it back from a specified environment. When creating and training a model, there are affordances in the interface to choose among a group of pre-existing archetype or template models. These models in the initial iteration of PredictChain are:
\begin{enumerate*}
    \item Multi-layered perceptron (MLP) neural network~\cite{preceptrons},
    \item Recurrent neural network (RNN) ~\cite{RNN},
    \item Long short-term memory neural network (LSTM) ~\cite{LSTM}, and
    \item Gated recurrent unit (GRU) neural network~\cite{GRU}.
\end{enumerate*}

Each of these models has a \textit{model\_complexity} attribute, which is a simple float value designed to give users a general idea of how performant a model can be once trained and serves as a method of calculating the cost of using or training that given model. The attribute itself is calculated using the size of the network and a linear multiplier to account for more complex model architectures. The complexity is higher for models like GRUs or LSTMs as they are more complex and often better-performing models~\cite{recurrentModeling}. Our RNN models can predict several time steps, but their quality often degrades when a lag between evidence and prediction is introduced~\cite{weightGuessing}. For these networks and our MLPs, the complexity is lower, which gives the desired effect of faster, simpler models being cheaper than the slower, more complex models without heavy calculations. The interface abstracts most of the complexities of training, querying, and evaluating these models. The only difference between them is the inclusion of several optional parameters.

\subsection{Tokenomics for Resource Sharing}

As the scale and power of machine learning models grow, the resources required to train them also grow. To get useful results from these models, users often have to pay large, centralized organizations, and they need to be incentivized to provide good datasets or model parameters. PredictChain changes this paradigm by incentivizing the thoughtful creation of useful and accessible models and datasets with a transparent pricing scheme. 
The underlying blockchain acts as a secure and transparent ledger that facilitates transactions between market participants, enabling them to exchange resources in a decentralized manner. 
The PredictChain marketplace allows individuals and organizations to share their data and computational resources, which can then be used to train machine learning models collaboratively. 
Tokenomics is crucial in incentivizing participation, rewarding contributors, and facilitating efficient transactions. 

In PredictChain, the medium of exchange is ALGO, which helps to simplify the transaction process as users will not be required to pay into yet another token hoping that its value does not fluctuate. We expect that users will feel more secure, as the lack of a separate token would make ``rug-pull'' deceptions much more difficult to execute. Our usage of ALGO also simplifies paying into or out of PredictChain.
In this model, participants can contribute their unused computing resources to the marketplace by hosting a node or submitting datasets that would increase the model performance, and in exchange, they receive ALGOs that represent their contribution. 
These tokens can then be used to access computing resources from other participants in the marketplace, purchase machine learning models trained on the data contributed by other participants, or be used as a medium of exchange anywhere in the Algorand ecosystem.

In PredictChain, the models and the datasets are publicly accessible to other users via IPFS. The rewards, given as $R$ microALGOs, for uploading datasets or training models are calculated using a public pre-determined equations. $ds\_size$ is the size of the dataset in bytes, $accuracy$ is the model's accuracy given the validation dataset or a real-world event, and the $*\_mult$ values are arbitrary values that the oracle sets to adjust the price as needed due to prevailing market conditions. Each calculation has its own value to balance out the calculations.  A complete log of these adjustments is stored on the blockchain.

To incentivize the good-faith hosting and maintenance of a PredictChain, those who host the hardware for the network will receive a small portion of each transaction.  The oracle can determine this rate, similar to the $mult$ values for the rewards. The results of every transaction would again be stored on the blockchain. Clients could then examine the history of any given oracle and determine if they offer fair transaction fees.  Given enough oracles hosting the network, this would allow clients great freedom in choosing the most competitive oracle.

\begin{equation}
\text{Dataset Usage and Querying Reward} = \lfloor ds\_size * mult * accuracy \rfloor
\label{eq:dataset-reward}
\end{equation}

\begin{equation}
\text{Model Training Reward} = \lfloor mult * accuracy \rfloor.
\label{eq:training-reward}
\end{equation}

The exact value of $mult$ would vary across each implementation of PredictChain.  The following are several example values and rewards across hypothetical models and datasets.  All values are in MicroALGO.

In the following two tables, the values in the first row represent the values of $mult$, the constant value used in reward calculations. Their variance allows flexibility when adjusting to fluctuating ALGO prices of other market pressures. The values in the first column represent the accuracy of the rewarded models or those trained on a given dataset. The values in the body of the graph are the reward values for model trainers or dataset uploaders in MicroALGO.



\begin{table}[H]
    \begin{minipage}{0.5\textwidth}
        \centering
        \begin{tabular}{|c|c|c|c|}
            \hline
            & \multicolumn{3}{c|}{\textit{mult}}\\
            \hline
            \multirow{6}{*}{accuracy} & & $\mathbf{10^7}$ & $\mathbf{3 \times 10^7}$\\
            \cline{2-4}
            & \textbf{0.3} & $3 \times 10^6$ & $9 \times 10^6$\\
            \cline{2-4}
            & \textbf{0.5} & $5 \times 10^6$ & $1.5 \times 10^7$\\
            \cline{2-4}
            & \textbf{0.7} & $7 \times 10^6$ & $2.1 \times 10^7$\\
            \cline{2-4}
            & \textbf{0.9} & $9 \times 10^6$ & $2.7 \times 10^7$\\
            \cline{2-4}
            & \textbf{0.99} & $9.9 \times 10^6$ & $2.97 \times 10^7$\\
            \hline
        \end{tabular}
        \caption{{Training Reward Example}}
        \label{tab:training-reward-example}
    \end{minipage}
    \begin{minipage}{0.5\textwidth}
        \centering
        \begin{tabular}{|c|c|c|c|}
            \hline
            $ds\_size=5 MB$ & \multicolumn{3}{c|}{\textit{mult}}\\
            \hline
            \multirow{6}{*}{accuracy} & & $\mathbf{2}$ & $\mathbf{6}$\\
            \cline{2-4}
            & \textbf{0.3} & $3 \times 10^6$ & $9 \times 10^6$\\
            \cline{2-4}
            & \textbf{0.5} & $5 \times 10^6$ & $1.5 \times 10^7$\\
            \cline{2-4}
            & \textbf{0.7} & $7 \times 10^6$ & $2.1 \times 10^7$\\
            \cline{2-4}
            & \textbf{0.9} & $9 \times 10^6$ & $2.7 \times 10^7$\\
            \cline{2-4}
            & \textbf{0.99} & $9.9 \times 10^6$ & $2.97 \times 10^7$\\
            \hline
        \end{tabular}
        \caption{{Database Upload Reward Example}}
        \label{tab:db-upload-reward}
    \end{minipage}
\end{table}

Note: The lower $mult$ values in the dataset reward calculations \ref{tab:db-upload-reward} is due to the size of the dataset.  When the $5 \times 10^6$ bytes of the 5MB dataset are factored into the calculation \ref{eq:dataset-reward}, the rewards for the two calculations \ref{eq:training-reward} \ref{eq:dataset-reward} are the same.

\subsection{Client}
The client is a middleman between the user and the blockchain. It is run as a server, serving user interface content to the user, taking in requests from the user, and parsing those requests into a form suitable for both the blockchain and the oracle.  It is separate from the oracle, being independent of any oracle implementation.  Additionally, the client constantly polls for updates from oracle through the blockchain and its monitor. These updates are queued and sent to the user interface upon request, allowing the user to interact with the blockchain and see the important updates that come from it.
Furthermore, the client sets the user's impression of PredictChain. For example, the homepage discusses our mission, how we differ from our competitors (such as the Shareable Updateable Sum framework by Harris and Wagonner~\cite{harris2020leveraging,harris2020analysis,sharingModels}, and example model sets we provide (LSTM, GRU, MLP, etc.). Users can create an account or log in to a pre-existing account via the client, and then use PredictChain to its fullest functionality, such as adding datasets, checking prices for model inferencing, or querying models with provided datasets. 


\subsection{Software and Libraries Used}

%
%
This work described in this paper is primarily built in Python, using the Algorand Software Development Kit (SDK)\footnote{\url{https://developer.algorand.org}}. The SDK makes interacting with the Algorand blockchain very straightforward, i.e., through the tools provided by this library, we can easily read and write transactions from the Algorand blockchain.

We also used data science libraries such as Pandas\footnote{\url{https://pandas.pydata.org}} and pyTorch\footnote{\url{https://pytorch.org}}. These libraries encapsulate many of the complexities of data preparation and model training. 
%
Flask\footnote{\url{https://flask.palletsprojects.com}} is also an important part of PredictChain's implementation. We used Flask to allow the client and oracle nodes to function as servers. The client would take in requests from the user and send out requests to the oracle. The oracle would then take in those requests and issue responses. We chose to use Flask in client-oracle communication to reduce the number of trivial transactions that would otherwise be made. For example, it would not benefit the accessibility or transparency of PredictChain greatly if the exchanged transactions were dominated by simple '\textit{what is the price of XXX?}' requests.
%
%
Additionally, the front-end client utilizes the React framework\footnote{\url{https://react.dev}} and Firebase\footnote{\url{https://firebase.google.com}}. React was useful to us as it streamlined the process of making dynamic, modular code for the web interface. Firebase was invaluable for handling administrative tasks such as keeping track of registered users and their associated metadata within a given node instance.
%
%
For the recommended configuration of PredictChain, we use Redis\footnote{\url{https://redis.io}} as well. Redis helps to provide a reliable store for our metadata about models and datasets in a simple, persistent manner.




\section{Evaluation}


\subsection{Transactions}

The usage of transactions is central to the communications protocol of PredictChain.  Algorand SDK allowed us to encode notes when sending transactions, which enabled us to ensure the protocol functions correctly.
Specifically, the monitors are classes inside the client and oracle that listen for any transactions with their node address as the recipient. This listening uses the Algorand indexer class and constant polling for new transactions. However, we noticed that this monitor would sometimes skip or duplicate incoming transactions. Therefore, we constantly evaluated the monitor's performance, which was done by programmatically sending one or more transactions to the client or oracle address and checking to see how the monitor handled them. We evaluated the monitor by comparing the unique transaction ids of the sent transactions to those processed. For example, we constantly update the minimum timestamp the indexer can look for transactions after, and we also keep a registry of the ids of all past transactions. This process helps to eliminate the duplicate transactions that were getting through.

\subsection{Models}

Another critical component of PredictChain is its usage of a variety of models. At present, all of these models are different types of neural networks. However, these networks are not created equally, each having different qualities, such as model parameters, architectures, and outputs. To evaluate the performance of these models, we ran several experiments on the models where their hyper-parameters and architectures were kept constant as they were tested. For an exemplary model evaluation, we measured the performance on our sample dataset, The University of California, Irvine's \textit{Dow Jones Index} data set~\cite{brown2013dynamic}.  This dataset has 16 different parameters detailing the attributes of a range of stocks from the first half of 2011. 
As for the models, they were all initialized with the parameters outlined in \Cref{tab:evalParams}.

\begin{table}[!htbp]
    \begin{center}
        \caption{{Model Evaluation Parameters}}
        \label{tab:evalParams}
        \bgroup
        \def\arraystretch{1.2}
        \begin{tabular}{|p{4cm}|p{1cm}|p{8cm}|}
            \hline
            \textbf{Parameter Name} & \textbf{Value} & \textbf{Description}\\
            \hline
            \textit{epochs} & 70 & The number of epochs that the model would train for\\
            \hline
            \textit{target\_attrib} & \textit{close} & The attribute that the model was trying to predict, this is the daily closing price of the stock\\
            \hline
            \textit{hidden\_dim} & 5 & The number of neurons in each hidden layer\\
            \hline
            \textit{num\_hidden\_layers} & 1 & The number of hidden layers\\
            \hline
            \textit{time\_lag} & 0 & The number of time steps that pass between the input window and the prediction\\
            \hline
            \textit{training\_lookback} & 10 & The number of time steps that recurrent models receive as input\\
            \hline
            \textit{sub\_split\_value} & 0 & The integer id of the stock to predict, in this case, it is \textit{\$AA} (Alcoa Corp)\\
            \hline
        \end{tabular}
        \egroup
    \end{center}
\end{table}

\begin{figure}[!htbp]
    \begin{minipage}{0.49\textwidth}
        \centering
        \includegraphics[width=\linewidth]{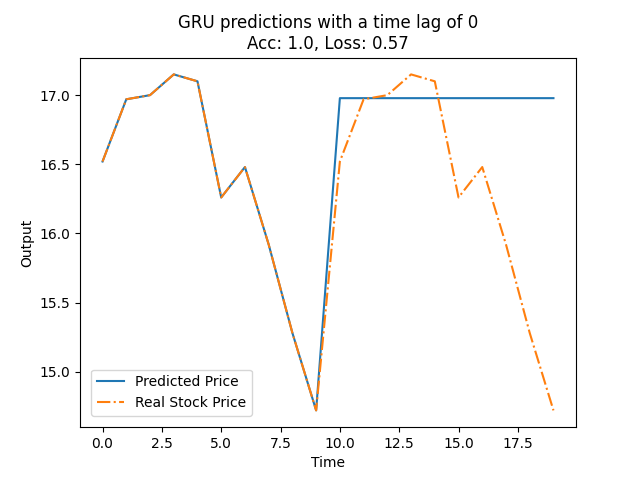}
        \caption{The results from our GRU model}\label{Fig:gruEval}
    \end{minipage}\hfill
    \begin{minipage}{0.49\textwidth}
        \centering
        \includegraphics[width=\linewidth]{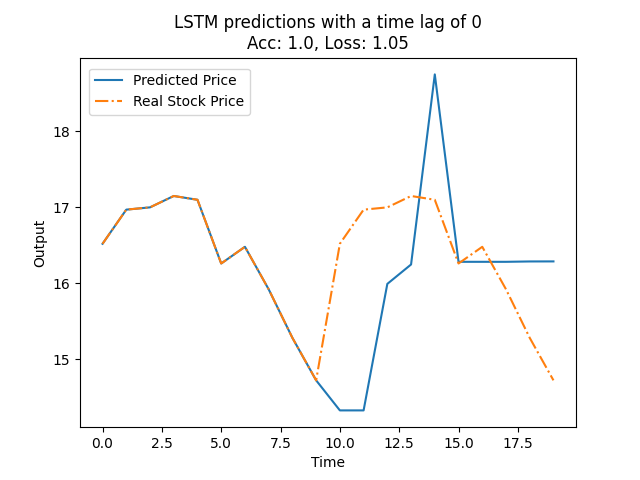}
        \caption{The results from our LSTM model}\label{Fig:lstmEval}
    \end{minipage}
    \begin{minipage}{0.49\textwidth}
        \centering
        \includegraphics[width=\linewidth]{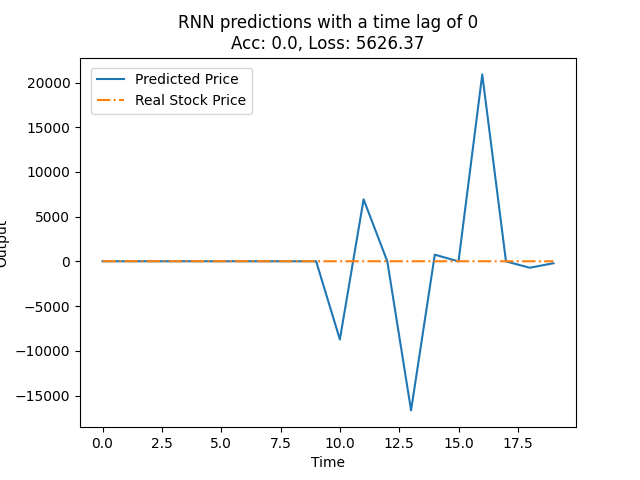}
        \caption{The results from our RNN model}\label{Fig:rnnEval}
    \end{minipage}\hfill
    \begin{minipage}{0.49\textwidth}
        \centering
        \includegraphics[width=\linewidth]{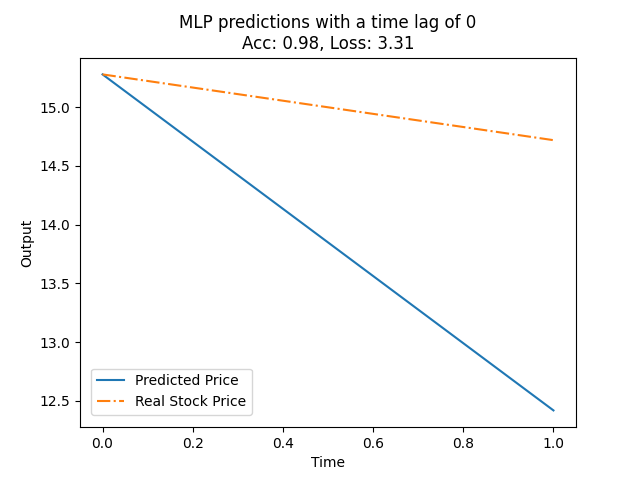}
        \caption{The results from our MLP model}\label{Fig:mlpEval}
    \end{minipage}
\end{figure}

We performed this test on all of our basic model structures, specifically our GRU, LSTM, RNN, and MLP models. In figures \ref{Fig:gruEval}, \ref{Fig:lstmEval}, \ref{Fig:rnnEval}, and \ref{Fig:mlpEval}, we show the input data and predictions for each model. We also show the final loss and the final accuracy. 



Each of these results reflects the individual strengths and weaknesses of each model. As is reflected in our \textit{model\_complexity} calculations, the GRU model is the most performant, with a loss of only 0.57, which is closely followed by the LSTM model, which still offers a good prediction, but is slightly less consistent overall. In our evaluation of our RNNs, we noticed a good example of the exploding gradients problem. Without the limits provided by the GRU and LSTM models, the predictions of the RNN become increasingly erratic. Finally, our MLP model does not suffer from an exploding gradient, but it does lack the recurrence of the other three models, only being able to output one day at a time.

\subsection{End-To-End Testing}

The final component of our testing was the end-to-end series of tests we performed. These tests gave us a holistic picture of how well PredictChain functioned and how well that functioning met our stated goals.
We started with front-end user interaction for all these tests and ended with the response to those actions.

\paragraph{Price Queries}
The three price query operations were the first and most straightforward operations we tested.
We performed this test by submitting query requests on the user interface, then tracing the functions called by that operation from the client to the oracle, then to the oracle getting the price, then finally back to the client. We performed this test for all three queries, testing different sets of valid and invalid inputs to see if they were handled properly.

\paragraph{Major Transactions}
The next set of tests we performed focused on our major transactions: \textit{Upload Dataset}, \textit{Train Model}, and \textit{Query Model}. These tests were similar to those for the price queries. As we performed these tests, we noted that, while the operations did complete successfully, the user feedback and results were ambiguous. Additionally, we realized that it might be confusing for the user to remember the exact names of the models and datasets to use them. To address these issues, we made several additions. To address the issue of users having to input the exact names, we added a dropdown feature instead of a test input. We now store a list of the system's current datasets and models, updated upon reloading the page or when a user submits a transaction. This process ensures that the list is always updated for better ease of use. To address the feedback issue, we added a feedback section that periodically pinged the client to see if any new response transactions came in from the oracle. If these transactions did come in, we would display the operation, the model or dataset name, and any extra data (like the result of a query) to the user. This way, the user would not manually check the transaction note on the Algorand block explorer.

By performing these tests, we evaluated the quality of PredictChain as a whole and made important additions to improve the user experience. Our transaction tests helped us to identify the issues that had previously existed and to verify that the current communication protocol was working properly. Our model tests helped us to confirm our previous assumptions about the nature of our various models and which scenarios they are most useful. Finally, our end-to-end testing confirmed the overall functioning of PredictChain and inspired us to make some valuable improvements to the user experience.



\section{Conclusion}

Through PredictChain, users can upload datasets for training predictive machine learning models, request that basic models be trained on any previously uploaded datasets, or submit queries to those trained models. Anybody participating in the PredictChain ecosystem can host a node with the necessary computing resources to make machine learning models available as a public good and benefit financially by naming a market-driven price for training these models on user-supplied datasets. PredictChain can enable various models, ranging from cheap, fast, and simple to more expensive, slower, and powerful. This will allow for various predictive abilities for both simple and complex needs. Furthermore, these models' past predictions are stored on the blockchain for public viewing and faster and free access by subsequent users.

As our work on PredictChain progresses, we recognize several opportunities for further enhancement. One key improvement involves providing users who upload datasets with greater flexibility in data preprocessing, potentially incorporating feature engineering techniques to optimize model performance. Additionally, expanding the repertoire of models offered on PredictChain beyond neural networks to include decision trees and statistical models based on Bayesian inference would enhance the marketplace's versatility. Furthermore, allowing users to customize the training process by pruning irrelevant attributes from datasets could lead to more tailored and efficient model training.
PredictChain can evolve into a more comprehensive and refined platform by implementing these future improvements while adhering to its core principles. Empowering users with greater control over their data and expanding the variety of models and datasets available will foster collaboration, innovation, and democratization of AI, propelling us towards new frontiers of knowledge and applications.

In conclusion, we have successfully developed PredictChain, a fully functional blockchain-based marketplace for predictive machine learning models. Users can leverage PredictChain to upload datasets for model training, request training on existing datasets, and submit queries to trained models. All past predictions are securely stored on the blockchain, promoting transparency and public accessibility.
With ongoing development and community involvement, PredictChain has the potential to revolutionize the AI landscape, shaping a future where accessibility, transparency, and shared intelligence are paramount.

\noindent\textbf{Resource Availability:} Our source code, demo, and comprehensive documentation are available in our GitHub repo at \url{https://github.com/AI-and-Blockchain/S23_PredictChain}.

\section*{Acknowledgements}
This work was supported by the Algorand Centres of Excellence program managed by the Algorand Foundation. Any opinions, findings, conclusions, or recommendations expressed in this material are those of the author(s) and do not necessarily reflect the views of the Algorand Foundation.






\bibliographystyle{unsrt}
\bibliography{main}






\end{document}